\documentclass[twoside,11pt]{article}
\usepackage{arxiv}

\title{Learning Feature Hierarchies with Centered Deep Boltzmann Machines}
\author{\name Gr\'egoire Montavon \email gmontavon@cs.tu-berlin.de\\
        \name Klaus-Robert M\"uller\thanks{Also at the Department of Brain and Cognitive Engineering, Korea University, Anam-dong, Seongbuk-gu, Seoul 136-713, Korea} \email klaus-robert.mueller@tu-berlin.de\\
        \addr Machine Learning Group\\
        Technische Unversit\"at Berlin\\
        Franklinstr. 28/29 \\ 10587 Berlin, Germany
}

\usepackage{array}
\usepackage{pstricks}
\usepackage{pst-plot}
\usepackage{pst-math}
\usepackage{amsmath}
\usepackage{amsfonts}
\usepackage{graphicx}
\usepackage[perpage,para,symbol*,bottom]{footmisc}

\usepackage{algpseudocode}
\usepackage{algorithmicx}
\usepackage{algorithm}

\bibpunct{(}{)}{;}{a}{,}{,}

\begin{document}

\maketitle

\newcommand{\diff}[2]{\frac{\partial #2}{\partial #1}}

\newcommand{\learningrate}{\eta}

\newcommand{\mnist}{{\small MNIST}}
\newcommand{\Mnist}{M{\small NIST}}
\newcommand{\sigm}{\text{sigm}}

\begin{abstract}%
Deep Boltzmann machines are in principle powerful models for extracting the hierarchical structure of data. Unfortunately, attempts to train layers jointly (without greedy layer-wise pretraining) have been largely unsuccessful. We propose a modification of the learning algorithm that initially recenters the output of the activation functions to zero. This modification leads to a better conditioned Hessian and thus makes learning easier. We test the algorithm on real data and demonstrate that our suggestion, the \emph{centered deep Boltzmann machine}, learns a hierarchy of increasingly abstract representations and a better generative model of data.
\end{abstract}



\section{Introduction}
\label{section-introduction}

Deep Boltzmann machines \citep[DBM,][]{salakhutdinov09} are in principle powerful models for extracting the hierarchical structure of data \citep{montavon12}. Unfortunately, attempts to train layers jointly (without greedy layer-wise pretraining) have been mostly unsuccessful. As we will argue later in greater detail, a possible reason for this could be that the mapping of net activities onto the sigmoid nonlinearities is not centered to zero by default.

In this paper, we propose to recenter the output of each unit to zero by rewriting the energy as a function of centered states $\xi = x - \beta$ where $\beta$ is an offset parameter. The reparameterization of the energy function leads to a better conditioned Hessian of the estimated model log-likelihood. The centered Boltzmann machine is easy to implement as the reparameterization leaves the associated Gibbs distribution invariant.

We train a centered deep Boltzmann machine on the \Mnist{} data set. Empirical results show that the centered DBM is able to learn a top-layer representation that contains useful discriminative features and to produce a good generative model of data. In addition, the centered DBM learns faster and is more stable than its non-centered counterpart.

\paragraph{Related work} The case for using centered nonlinearities has already been made by \citet{lecun98b} and \citet{glorot10} in the context of backpropagation networks, showing that the logistic function generally performs poorly compared to its centered counterpart, the hyperbolic tangent. The idea of centering was also proposed by \citet{tang11} in the context of restricted Boltzmann machines but was restricted to data centering.

\section{Centered Boltzmann Machines}
\label{section-centered}

In this section, we introduce the centered Boltzmann machine. In the following, the sigmoid function is defined as $\sigm(x) = \frac{e^x}{e^x+1}$, $x \sim \mathcal{B}(p)$ denotes that the variable $x$ is drawn randomly from a Bernouilli distribution of parameter $p$ and $\langle \cdot \rangle_{P}$ denotes the expectation operator with respect to a probability distribution $P$. All these operations apply element-wise to the input vector.

A Boltzmann machine is a network of $M_x$ interconnected binary units that associates to each state $x \in \{0,1\}^{M_x}$ the energy
\begin{align*}
E(x;\theta) = - x^\top W x - x^\top b
\end{align*}
where $\theta = \{W,b\}$ groups the model parameters. The matrix $W$ of size $M_x \times M_x$ is symmetric and contains the connection strength between units. The vector $b$ of size $M_x$ contains the biases associated to each unit. A probability is associated to each state according to the Gibbs distribution
\begin{align*}
p(x; \theta) = \frac{e^{-E(x;\theta)}}{\sum_{x} e^{-E(x;\theta)}}
\end{align*}
where the term in the denominator is the partition function that makes probabilities sum to one. For the centered Boltzmann machine, we rewrite the energy as a function of centered states
\begin{align*}
E(x;\theta) = - (x-\beta)^\top W (x-\beta) - (x-\beta)^\top b
\end{align*}
where $\theta = \{ W, b, \beta \}$ and where the vector $\beta$ contains the offsets associated to each unit of the network. Setting $\beta = \sigm(b_0)$ where $b_0$ is the initial bias enforces the initial centering of the Boltzmann machine. From these equations, we can derive the conditional probability
\begin{align*}
p(x_i = 1 | x_{-i}; \theta) = \sigm(b_i + \sum_{j \neq i} W_{ij}(x-\beta)_j)
\end{align*}
of each unit and the gradient of the model log-likelihood with respect to $W$ and $b$:
\begin{align*}
\diff{W}{}\langle \log p(x; \theta) \rangle_\text{data} &= \langle (x-\beta) (x-\beta)^\top \rangle_\text{data} -  \langle (x-\beta) (x-\beta)^\top \rangle_\text{model}\\
\diff{b}{}\langle \log p(x; \theta) \rangle_\text{data} &= \langle x-\beta \rangle_\text{data} - \langle x-\beta \rangle_\text{model}
\end{align*}

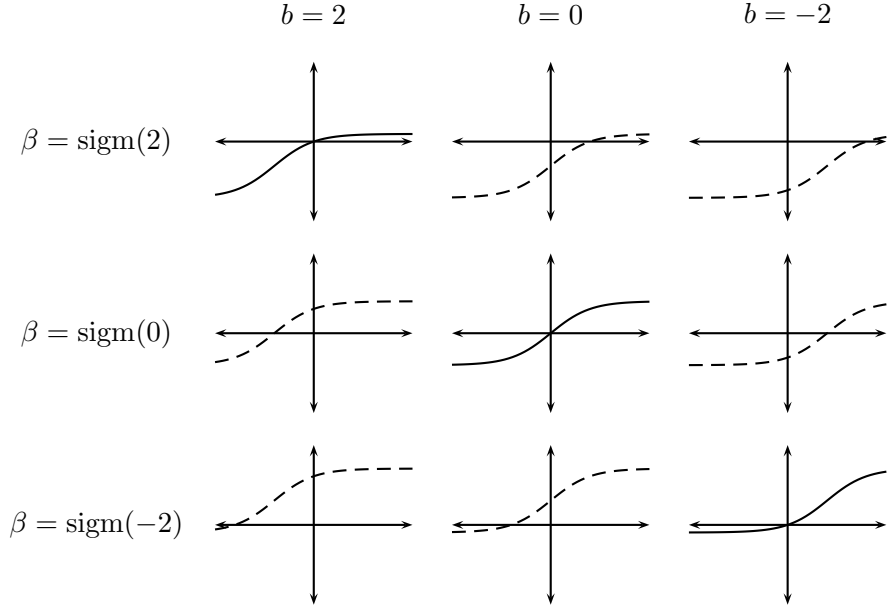
\begin{figure}
\centering
\newcommand{\plotsigm}[4]{\psline{<->}(-2.5,0)(2.5,0)\psline{<->}(0,-2.5)(0,2.5)
\psplot[linecolor=#3,linestyle=#4]{-2.5}{2.5}{x #1 sub TANH 0 #2 sub TANH sub}}
\psset{xunit=0.525cm,yunit=0.425cm}
\begin{pspicture*}(-14,-9)(9,10.5)
\rput(-11.5,-6){$\beta = \sigm(-2)$}
\rput(-11.5,0){$\beta = \sigm(0)$}
\rput(-11.5,6){$\beta = \sigm(2)$}
\rput(-6,10){$b = 2$}
\rput(0,10){$b = 0$}
\rput(6,10){$b = -2$}
\rput(-6,6){\plotsigm{-1}{-1}{black}{solid}}
\rput(-6,0){\plotsigm{-1}{0}{black}{dashed}}
\rput(-6,-6){\plotsigm{-1}{1}{black}{dashed}}
\rput(0,6){\plotsigm{0}{-1}{black}{dashed}}
\rput(0,0){\plotsigm{0}{0}{black}{solid}}
\rput(0,-6){\plotsigm{0}{1}{black}{dashed}}
\rput(6,6){\plotsigm{1}{-1}{black}{dashed}}
\rput(6,0){\plotsigm{1}{0}{black}{dashed}}
\rput(6,-6){\plotsigm{1}{1}{black}{solid}}
\end{pspicture*}
\caption{Example of sigmoids with different biases and offsets. The three non-dashed sigmoids are said to be \emph{centered} because they cross the origin. We show that centering sigmoids leads to a better conditioned Hessian.}
\label{figure-sigmoids}
\end{figure}

\subsection{Stability of the Centered Boltzmann Machine}

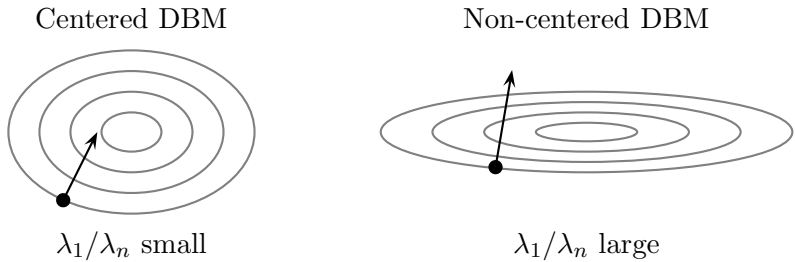
\begin{figure}
\centering
\psset{xunit=0.55cm,yunit=0.55cm,arrowscale=1.5}
\begin{pspicture}(0,-0.5)(20,5.5)
\rput(3,2.5){
	\psellipse[linecolor=gray](0,0)(0.75,0.5)
	\psellipse[linecolor=gray](0,0)(1.5,1)
	\psellipse[linecolor=gray](0,0)(2.25,1.5)
	\psellipse[linecolor=gray](0,0)(3,2)
	\psline{*->}(-1.65,-1.65)(-0.825,0)
	\rput(0,2.75){Centered DBM}
	\rput(0,-2.75){$\lambda_1/\lambda_n$ small}
}
\rput(14,2.5){
	\psellipse[linecolor=gray](0,0)(1.25,0.25)
	\psellipse[linecolor=gray](0,0)(2.5,0.5)
	\psellipse[linecolor=gray](0,0)(3.75,0.75)
	\psellipse[linecolor=gray](0,0)(5,1)
	\psline{*->}(-2.2,-0.85)(-1.8,1.5)
	\rput(0,2.75){Non-centered DBM}
	\rput(0,-2.75){$\lambda_1/\lambda_n$ large}
}
\end{pspicture}
\caption{Relation between the conditioning number $\lambda_1/\lambda_n$ and the shape of the optimization problem. Gradient descent is easier to achieve when the conditioning number is small.}
\label{figure-conditioning}
\end{figure}

In this section, we look at the stability of the underlying optimization problem. We argue that when the sigmoid is centered, the Hessian is better conditioned (see Figure \ref{figure-conditioning}), and therefore, the learning algorithm is more stable. We define $\xi$ as the centered state $\xi = x - \beta$. The derivative of the model log-likelihood with respect to the weight vector takes the form
\begin{align*}
\diff{W}{} \langle \log p(x; \theta) \rangle_\text{data} =  \langle \xi\xi^\top \rangle_\text{data} - \langle \xi\xi^\top \rangle_W
\end{align*}
where $\langle \cdot \rangle_W$ designates the expectation with respect to the probability distribution associated to a model of weight parameter $W$. Using the definition of the directional derivative, the second derivative with respect to a random direction $V$ (which is equal to the projected Hessian $\boldsymbol H V$) can be expressed as:
\begin{align*}
\boldsymbol H V &= \diff{V}{} \left( \diff{W}{} \langle \log p(x;W) \rangle_\text{data} \right)\\
&=\lim_{h \to 0} \frac1h \left( \diff{W}{}\langle \log p(x;W+h V) \rangle_\text{data} - \diff{W}{}\langle \log p(x;W) \rangle_\text{data} \right)\\
&= \lim_{h \to 0} \frac1h \left( (\langle \xi\xi^\top \rangle_{W+hV,\text{data}} - \langle \xi\xi^\top \rangle_{W+h V}) - ( \langle \xi\xi^\top \rangle_{W,\text{data}} - \langle \xi\xi^\top \rangle_{W}) \right)\\
&= \lim_{h \to 0} \frac1h \left( \langle \xi\xi^\top \rangle_{W+hV,\text{data}} - \langle \xi\xi^\top \rangle_{W,\text{data}}  \right) -  \lim_{h \to 0} \frac1h \left( \langle \xi\xi^\top \rangle_{W+h V} - \langle \xi\xi^\top \rangle_{W} \right)
\end{align*}
From the last line, we can see that the Hessian can be decomposed into a data-dependent term and a data-independent term. A remarkable fact is that in absence of hidden units, the data-dependent part of the Hessian is zero, because the model---and therefore, the perturbation of the model---have no influence on the states. The conditioning of the optimization problem can therefore be analyzed exclusively from the perspective of the model without even looking at the data. The data-dependent term is likely to be small even in the presence of hidden variables due to the sharp reduction of entropy caused by the clamping of visible units to data.

We can think of a well-conditioned model as a model for which a perturbation of the model parameter $W$ in any direction $V$ causes a well-behaved perturbation of state expectations $\langle \xi\xi^\top \rangle_W$.  \citet{pearlmutter94} showed that in a Boltzmann machine with no hidden units, the projected Hessian can be further reduced to
\begin{align}
\boldsymbol H V = \langle \xi\xi^\top \rangle_W \cdot \langle D \rangle_W - \langle \xi\xi^\top D \rangle_W \qquad \text{where} \quad D = \frac12 \xi^\top V \xi
\label{equation-pearlmutter}
\end{align}
thus, getting rid of the limit and leading to numerically more accurate estimates. \citet{lecun98b} showed that the stability of the optimization problem can be quantified by the \emph{conditioning number} defined as the ratio between the largest eigenvalue $\lambda_1$ and the smallest eigenvalue $\lambda_n$ of $\boldsymbol H$. A geometrical interpretation of the conditioning number is given in Figure \ref{figure-conditioning}. A low rank approximation of the Hessian can be obtained as
\begin{align}
\hat{\boldsymbol H} = \boldsymbol H (V_0 | \hdots | V_n) =  (  \boldsymbol H V_0 | \hdots |  \boldsymbol H V_n)
\label{equation-rproj}
\end{align}
where the columns of $(V_0 | \hdots | V_n)$ form a basis of independent unit vectors that projects the Hessian on a low-dimensional random subspace. The conditioning number can then be estimated by performing a singular value decomposition of the projected Hessian $\hat{\boldsymbol H}$ and taking the ratio between the largest and smallest resulting eigenvalues.

We estimate below the conditioning number $\lambda_1 / \lambda_n$ of a fully connected Boltzmann machine of $50$ units at initial state ($W = 0$) for different bias and offset parameters $b$ and $\beta$ using Equation \ref{equation-pearlmutter} and \ref{equation-rproj}:

\begin{center}
\begin{tabular}{c|ccc}
$\lambda_1 / \lambda_n $ & $b = 2$ & $b = 0$ & $b = -2$ \\\hline
$\beta = \sigm(2)$ & \textbf{2.26} & 21.97 & 839.59\\
$\beta = \sigm(0)$ & 83.43 & \textbf{2.75} & 95.57 \\
$\beta =\sigm(-2)$ & 866.00 & 22.95 & \textbf{2.24}\\
\end{tabular}
\end{center}

These numerical estimates clearly exhibit the better conditioning occuring when the sigmoid is centered. The more than 100-fold factor between the conditioning number of non-centered and centered Boltzmann machines is striking.

\subsection{Centered Deep Boltzmann Machines}

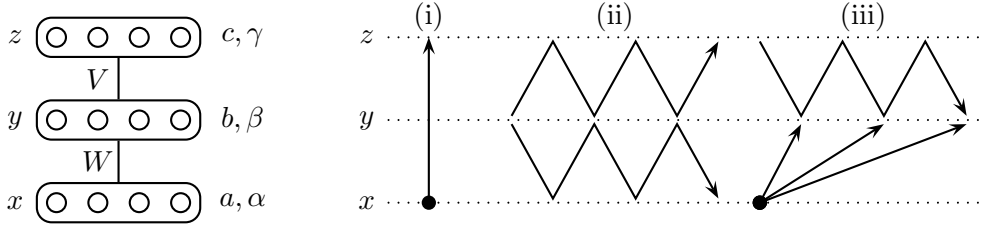
\begin{figure}
\centering
\psset{xunit=0.55cm,yunit=0.55cm,arrowscale=1.5}
\begin{pspicture}(-7.5,-0.5)(16.5,5)
\pscircle(-6.0,0){0.14} \pscircle(-5.0,0){0.14} \pscircle(-4.0,0){0.14} \pscircle(-3.0,0){0.14} \psframe[framearc=0.75](-6.5,-0.5)(-2.5,0.5) \rput(-7,0){$x$} \rput(-1.5,0){$a,\alpha$}
\pscircle(-6.0,2){0.14} \pscircle(-5.0,2){0.14} \pscircle(-4.0,2){0.14} \pscircle(-3.0,2){0.14} \psframe[framearc=0.75](-6.5, 1.5)(-2.5,2.5) \rput(-7,2){$y$} \rput(-1.5,2){$b,\beta$}
\pscircle(-6.0,4){0.14} \pscircle(-5.0,4){0.14} \pscircle(-4.0,4){0.14} \pscircle(-3.0,4){0.14} \psframe[framearc=0.75](-6.5, 3.5)(-2.5,4.5) \rput(-7,4){$z$} \rput(-1.5,4){$c,\gamma$}
\psline{-}(-4.5,0.5)(-4.5,1.5) \rput(-5,1){$W$}
\psline{-}(-4.5,2.5)(-4.5,3.5) \rput(-5,3){$V$}

\rput(2,0){
\psline[linestyle=dotted](0,0)(14.5,0) \rput(-0.5,0){$x$}
\psline[linestyle=dotted](0,2)(14.5,2) \rput(-0.5,2){$y$}
\psline[linestyle=dotted](0,4)(14.5,4) \rput(-0.5,4){$z$}
\psline{*->}(1,0)(1,4)                                                                         \rput( 1.0,4.5){(i)}
\psline{->}(3,2.1)(4,3.9)(5,2.1)(6,3.9)(7,2.1)(8,3.9)      \psline{->}(3,1.9)(4,0.1)(5,1.9)(6,0.1)(7,1.9)(8,0.1)       \rput( 5.5,4.5){(ii)}
\psline{->}(9,3.9)(10,2.1)(11,3.9)(12,2.1)(13,3.9)(14,2.1)
\psline{*->}(9,0)(10,1.9) \psline{*->}(9,0)(12,1.9) \psline{*->}(9,0)(14,1.9)  \rput(11.5,4.5){(iii)}
}
\end{pspicture}
\caption{On the left, diagram of a two-layer deep Boltzmann machine along with its parameters. On the right, different sampling methods: (i) a feed-forward pass on the network starting from a data point, (ii) the path followed by the alternate Gibbs sampler and (iii) the path followed by the alternate Gibbs sampler when the input is clamped to data.}
\label{figure-dbm}
\end{figure}

For technical and practical reasons, it is common to introduce a structure to the Boltzmann machine by restricting the connections between its units. A typical structure is the deep Boltzmann machine \citep[DBM,][]{salakhutdinov09} in which units are organized in a deep layered architecture. The layered structure of the DBM has two advantages: first, it gives a specific role to units at each layer so that we can easily build top layer kernels that exploit the hierarchical structure of data. Second, the layered structure of the DBM can be folded into a bipartite graph from which it is easy to derive an efficient alternate Gibbs sampler. In the case of the two-layer deep Boltzmann machine shown in Figure \ref{figure-dbm}, the energy function associated to each state $(x,y,z) \in \{0,1\}^{M_x+M_y+M_z}$ takes the form
\begin{align*}
E(x,y,z;\theta) = &- (y-\beta)^\top W (x-\alpha) - (z-\gamma)^\top V (y-\beta)\\
                       &- (x-\alpha)^\top a - (y-\beta)^\top b - (z-\gamma)^\top c
\end{align*}
where $\theta = \{W,V,a,b,c,\alpha,\beta,\gamma\}$ groups the model parameters. Data-\emph{independent} states can be sampled using the following alternate Gibbs sampler:
\begin{gather}
\label{equation-dbmgibbssampler}
\{ x \sim \mathcal{B}(\sigm(W^\top (y-\beta) + a)) \quad ; \quad z \sim \mathcal{B}(\sigm(V (y-\beta) + c)) \}\\
y \sim \mathcal{B}(\sigm(W (x- \alpha) + V^\top (z - \gamma) + b)) .
\end{gather}
The same Gibbs sampler can be used for sampling data-\emph{dependent} states at the difference that the input units $x$ are clamped to the data. We show below a basic algorithm based on persistent contrastive divergence for training a two-layer centered DBM:

\begin{center}
\fbox{
\begin{minipage}[t]{0.9\textwidth}
\vskip 1mm
\textbf{Basic algorithm for training a 2-layer centered DBM:}
\vskip 3mm
\small
\begin{algorithmic}
\State $W, V = 0, 0$
\State $a, b, c =\sigm^{-1}(\langle x \rangle_\text{data}), b_0, c_0$
\State $\alpha, \beta, \gamma =\sigm(a), \sigm(b), \sigm(c)$
\State initialize free particle $(x_m,y_m,z_m) = (\alpha,\beta,\gamma)$
\Loop
\State initialize data particle $(x_d,y_d,z_d) = (\text{pick}(\text{data}),\beta,\gamma)$
\Loop
\State $y_d \sim \mathcal{B}(\sigm(W (x_d-\alpha) + V^\top (z_d-\gamma) + b))$
\State $z_d \sim \mathcal{B}(\sigm(V (y_d-\beta) + c))$
\EndLoop
\vskip 1mm
\State $y_m \sim \mathcal{B}(\sigm(W (x_m-\alpha) + V^\top (z_m- \gamma) + b))$
\State $x_m \sim \mathcal{B}(\sigm(W^\top (y_m-\beta) + a))$
\State $z_m \sim \mathcal{B}(\sigm(V (y_m-\beta) + c))$
\vskip 1mm
\State $W = W + \learningrate \cdot [(y_d-\beta)(x_d-\alpha)^\top - (y_m-\beta)(x_m-\alpha)^\top]$
\State $V = V + \learningrate \cdot [(z_d-\gamma)(y_d-\beta)^\top -  (z_m-\gamma)(y_m-\beta)^\top]$
\vskip 1mm
\State $a = a + \learningrate \cdot (x_d - x_m)$
\State $b = b + \learningrate \cdot  (y_d - y_m)$
\State $c = c + \learningrate \cdot (z_d - z_m)$
\vskip 1mm
\EndLoop
\end{algorithmic}
\end{minipage}
}
\end{center}

\section{Discriminative Analysis}
\label{section-discriminative}

We present the method introduced by \citet{montavon11} that measures how the representation evolves layer after layer in a deep network. It is based on the theoretical insight that the projection of the input distribution onto the hidden units of each layer provides a function space that can be thought of as a representation or a feature extractor.

The method aims to characterize this function space by constructing a kernel for each layer that approximates the implicit transfer function between the input and the layer and measuring how much these kernels ``match'' the task of interest. The approach is theoretically motivated by the work of \citet{braun08} showing that projections on the leading components of the implicit kernel feature map \citep{schoelkopf98} obtained with a finite and typically small number of samples $n$ are close with essentially multiplicative errors to their asymptotic counterparts. In the following lines, we describe the principal steps of the analysis:

Let $X$ and $T$ be two matrices of $n$ rows representing respectively the inputs and labels of a data set of $n$ samples. Let
\begin{align*}
f : x \mapsto f_L \circ \dots \circ f_1 (x)
\end{align*}
be a deep network of $L$ layers. We build a hierarchy of increasingly ``deep'' kernels
\begin{align*}
k_{0,\sigma}(x,x') &= \kappa_\sigma(x,x')\\
k_{1,\sigma}(x,x') &= \kappa_\sigma(f_1(x),f_1(x'))\\
& ~\: \vdots\\
k_{L,\sigma}(x,x') &= \kappa_\sigma(f_L \circ \dots \circ f_1(x),f_L \circ \dots \circ f_1(x'))
\end{align*}
that subsume the mapping performed by more and more layers of the deep network and where $\kappa_\sigma$ is an RBF kernel of scale $\sigma$. For each kernel $k_{l,\sigma}$, we can compute the empirical kernel $K_{l,\sigma}$ of size $n \times n$ and its eigenvectors $u_{l,\sigma}^1,\dots, u_{l,\sigma}^n$ sorted by decreasing magnitude of their respective eigenvalues $\lambda_{l,\sigma}^1,\dots,\lambda_{l,\sigma}^n$. 

We measure how good a representation is with respect to a certain task by measuring whether the task is contained within the leading principal components of the representation. The matrix
\begin{align*}
U_{l,\sigma}^d = (u_{l,\sigma}^1  \mid \dots  \mid  u_{l,\sigma}^d)
\end{align*}
spans the $d$ leading kernel principal components of empirical kernel. The error is obtained as the residuals of the projection of the labels $T$ on the $d$ leading components of the mapped distribution:
\begin{align*}
e_T(l,d,\sigma) = || T - U_{l,\sigma}^d {U_{l,\sigma}^d}^{\!\!\!\top} T ||_F^2
\end{align*}

Curves $(e(l,0,\sigma),\dots,e(l,d,\sigma))$ represent how well the task can be solved as we add more and more principal components of the data distribution. These curves can be interpreted as learning curves as the regularization imposed by the rank of the kernel feature space determines the number of samples that are necessary in order to train the model effectively. Therefore, the number of observed kernel principal components $d$ closely relates to the amount of label information given to the learning machine. Small values for $d$ cover the ``one-shot'' learning regime where the model is asked to generalize from very few observations. On the other hand, large values for $d$ cover the other extreme case where label information is abundant, and where the representation has to be rich enough in order to encode any subtle variation of the learning problem. For practical purposes, these curves can be reduced as follows:
\begin{align}
e_T(l,d) &= \min_\sigma e_T(l,d,\sigma)
\label{equation-kpca}\\
e_T(l) &= \frac1n \sum_{d=1}^{n} e_T(l,d)
\label{equation-auc}
\end{align}
These compact measures of how well layer $l$ represent $T$ make it easier to compare the layer-wise evolution of the representation for different architectures.

\section{Generative Analysis}
\label{section-generative}

Here, we present an analysis that estimates the likelihood of the learned Boltzmann machine \citep{salakhutdinov10} based on annealed importance sampling \citep[AIS,][]{neal01}. We describe here the basic analysis. \citet{salakhutdinov10} introduced more elaborate procedures for particular types of Boltzmann machines such as restricted, semi-restricted and deep Boltzmann machines.

A deep Boltzmann machine associates to each input $x$ a probability
\begin{align*}
p(x;\theta) = \frac{\Psi(\theta,x)}{Z(\theta)}
\end{align*}
\begin{align*}
\text{where} \quad  \Psi(\theta,x) &= \sum_{y,z} p^\star(x,y,z;\theta) \\
 Z(\theta) &= \sum_{x,y,z} p^\star(x,y,z;\theta)
\end{align*}
and where $p^\star(x,y,z;\theta) = e^{-E(x,y,z;\theta)}$ is the unnormalized probability of state ($x, y, z$). Computing $\Psi(\theta,x)$ and $Z(\theta)$ analytically is intractable because of the exponential number of elements involved in the sum. Let us rewrite the ratio of partition functions as follows:
\begin{align}
\label{equation-decomposition}
p(x;\theta) &= \frac{\Psi(\theta,x)}{Z(\theta)} = 
\frac{  \frac{\Psi(\theta,x)}{\Psi(0,x)}  } {  \frac{Z(\theta)}{Z(0)}  } \cdot
 \frac{\Psi(0,x)}{Z(0)} 
\end{align}
It can be first noticed that the ratio of base-rate partition functions ($\theta = 0$) is easy to compute as $\theta = 0$ makes units independent. It has the analytical form
\begin{align}
\label{equation-baserate}
\frac{\Psi(0,x)}{Z(0)}  = \frac1{2^{M_x}}.
\end{align}
The two other ratios in Equation \ref{equation-decomposition} can be estimated using annealed importance sampling. The annealed importance sampling method proceeds as follows:
\begin{center}
\fbox{
\begin{minipage}[t]{0.9\textwidth}
\vskip 2mm
\textbf{Annealed importance sampling:}
\begin{enumerate}
\item Generate a sequence of states $\xi_1,\dots,\xi_T$ using a sequence of transition operators $\mathcal{T}(\xi,\xi'; \theta_0), \dots, \mathcal{T}(\xi,\xi'; \theta_K)$ that leave $p(\xi)$ invariant, that is,
\begin{itemize}
\item Draw $\xi_0$ from the base model (e.g. a random vector of zero and ones)
\item Draw $\xi_1$ given $\xi_0$ using $\mathcal{T}(\xi,\xi'; \theta_1)$
\item $\dots$
\item Draw $\xi_K$ given $\xi_{K-1}$ using $\mathcal{T}(\xi,\xi'; \theta_K)$
\end{itemize}
\item Compute the importance weight
\begin{align*}
\omega_\text{AIS} =
\frac{p^\star(\xi_1;\theta_1)}{p^\star(\xi_1;\theta_0)} \cdot
\frac{p^\star(\xi_2;\theta_2)}{p^\star(\xi_2;\theta_1)} \cdot \dots \cdot
\frac{p^\star(\xi_K;\theta_K)}{p^\star(\xi_K;\theta_{K-1})} 
\end{align*}
\end{enumerate}
\vskip 2mm
\end{minipage}
}
\end{center}

It can be shown that if the sequence of models $\theta_0,\theta_1,\dots,\theta_K$ where $\theta_0 = 0$ and $\theta_K = \theta$ evolves slowly enough, the importance weight obtained with the annealed importance sampling procedure is an estimate for the ratio between the partition function of the model $\theta$ and the partition function of the base rate model.

In our case, $\xi$ denotes the state $(x,y,z)$ of the DBM and the transition operator $\mathcal{T}(\xi,\xi'; \theta)$ is the alternate Gibbs sampler defined in Equation \ref{equation-dbmgibbssampler}. We can now compute the two ratios of partition functions of Equation \ref{equation-decomposition} as
\begin{align}
\label{equation-approxratio}
\frac{Z(\theta)}{Z(0)} \approx \text{E} [ \omega_\text{AIS} ]
\quad \text{and} \quad
\frac{\Psi(\theta,x)}{\Psi(0,x)} \approx \text{E} [ \nu_\text{AIS}(x) ]
\end{align}
where $\omega_\text{AIS}$ is the importance weight resulting from the annealing process with the freely running Gibbs sampler and $\nu_\text{AIS}$ is the importance weight resulting from the annealing with input units clamped to the data point. Substituting Equation \ref{equation-baserate} and \ref{equation-approxratio} into Equation \ref{equation-decomposition}, we obtain 
\begin{align*}
p(x;\theta) \approx \frac{\text{E} [\nu_\text{AIS}(x)] }{ \text{E} [\omega_\text{AIS}] } \cdot \frac{1}{2^{M_x}}
\end{align*}
and therefore, the log-likelihood of the model is
\begin{align}
\label{equation-ais-loglikelihood}
\text{E}_X[\log(p(x;\theta))] \approx \text{E}_X [ \log \text{E} [\nu_\text{AIS}(x)] ] - \log \text{E} [\omega_\text{AIS}] - M_x \log(2).
\end{align}
Generally, computing an average of the importance weight $\nu_\text{AIS}$ for each data point $x$ can take a long time. In practice, we can use an approximation to this computation where the estimate is computed with a single AIS run for each point. In that case, it follows from Jensen's inequality that
\begin{align}
\label{equation-ais-jensen}
\text{E}_X [ \log \nu_\text{AIS}(x) ] - \log \text{E} [\omega_\text{AIS}] 
\leq
\text{E}_X [ \log \text{E} [\nu_\text{AIS}(x)] ] - \log \text{E} [\omega_\text{AIS}] .
\end{align}
Consequently, this approximation tends to produce slightly pessimistic estimates of the model log-likelihood, however the variance of $\nu_\text{AIS}$ is low compared to the variance of $\omega_\text{AIS}$ because the clamping of visible units to data points sharply reduces the diversity of AIS runs. We find that this approximation is sufficiently accurate for the purpose of this paper, that is, demonstrating the importance of centering deep Boltzmann machines.

\section{Experimental Setup}
\label{section-methodology}

In this section, we describe the different parameters used to train the deep Boltzmann machines and to perform the discriminative and generative analysis. These parameters correspond to reasonable choices, most of which have been validated by previous research work.

\paragraph{Architecture} We consider two-layer deep Boltzmann machines made of $784$ input units, $400$ intermediate units and $100$ top units.
The initial biases and offsets for visible units are set to $a_0 = \sigm^{-1}(\langle x \rangle_\text{data})$ and $\alpha = \sigm(a)$. We consider different initial biases ($b_0,c_0 = -2$, $b_0,c_0 = 0$ and $b_0,c_0 = 2$) and offsets ($\beta,\gamma = \sigm(-2)$, $\beta,\gamma = \sigm(0)$ and $\beta,\gamma = \sigm(2)$) for the hidden units. These offsets and initial biases correspond to the sigmoids plotted in Figure \ref{figure-sigmoids}.

\paragraph{Data} We train the DBMs on a binary version of the \Mnist{} handwritten digits data set where the activation threshold is set to 0.5 (medium gray). The \Mnist{} training set consists of 60,000 samples. Each sample is a binary image of size $28 \times 28$ representing a handwritten digit and is fed to the DBM as a 784-dimensional binary vector.

\paragraph{Inference} We use persistent contrastive divergence \citep{tieleman08} to train the network and keep track of $25$ free particles in background of the learning procedure. We use a Gibbs sampling estimation to collect \emph{both} the data-independent and data-dependent statistics. The rationale for this is that the more classical mean field estimation of data statistics \citep{salakhutdinov09} tends to artificially drive the DBM to sparsity due to the convex/concave shape of the sigmoid function. At each step of the learning procedure, we run $5$ iterations of the alternate Gibbs sampler for collecting the data-dependent statistics and one iteration for updating the data-independent statistics.

\paragraph{Learning} We use a stochastic gradient descent on the approximate log-likelihood with minibatches of size $25$ and a learning rate $\eta = 0.0005$ for each layer. For practical purposes, the minibatch size is set equivalent to the number of particles for persistent contrastive divergence \citep{hinton10}. We consider models trained for $10^0$, $10^{0.5}$, $10^1$, $10^{1.5}$ and $10^2$ epochs.

\paragraph{Model averaging} We use a variant of averaged stochastic gradient descent \citep{polyak92,tieleman09,xu11} for reducing the parameter noise. We compute at each step $k$ the new parameter estimate $\theta_\text{avg} \leftarrow \frac{k_c}{k+k_c} \cdot \theta + \frac{k}{k+k_c} \cdot \theta_\text{avg}$ with $k_c=10$ in order to only remember the last $10\%$ of the training procedure.

\paragraph{Discriminative analysis} The analysis is performed on a subset of $500$ samples drawn randomly from the \Mnist{} test set. Representations at each layer are built by running a Gibbs sampler for 100 iterations with the input clamped to data and taking the mean activation of each unit. Discriminative performance is measured as the projection residuals of the labels (see Equation \ref{equation-kpca}) and the area under the error curve (see Equation \ref{equation-auc}). Results are produced with candidate scale parameters of the Gaussian kernel $\sigma^2 = 1$, $10$, $100$, $1000$ and $10000$.

\paragraph{Generative analysis} The generative analysis is performed on a subset of $500$ samples drawn randomly from the \Mnist{} test set. Generative performance is measured as the estimated log-likelihood of the model given the test data (see Equation \ref{equation-ais-loglikelihood}). We estimate the partition function $Z(\theta)/Z(0)$ using $500$ AIS runs. We estimate each $500$ partition functions  $\Psi(\theta,x)/\Psi(0,x)$ using a single AIS run. Each AIS run has length $K = 2500$ where model parameter at the $k^\text{th}$ step of the annealing process is defined as $\theta_k = 1-(1-k\theta/K)^2$. This sequence of parameters implies that annealing starts with large parameter updates and finishes with very small updates.

\section{Results}
\label{section-results}

\begin{figure}
\centering
\begin{tabular}{cc}
~~~~Discriminative analysis of $k_2(x,x')$  & ~~~~Layer-wise evolution of the representation\\
\includegraphics[scale=0.7,trim=10 0 0 0]{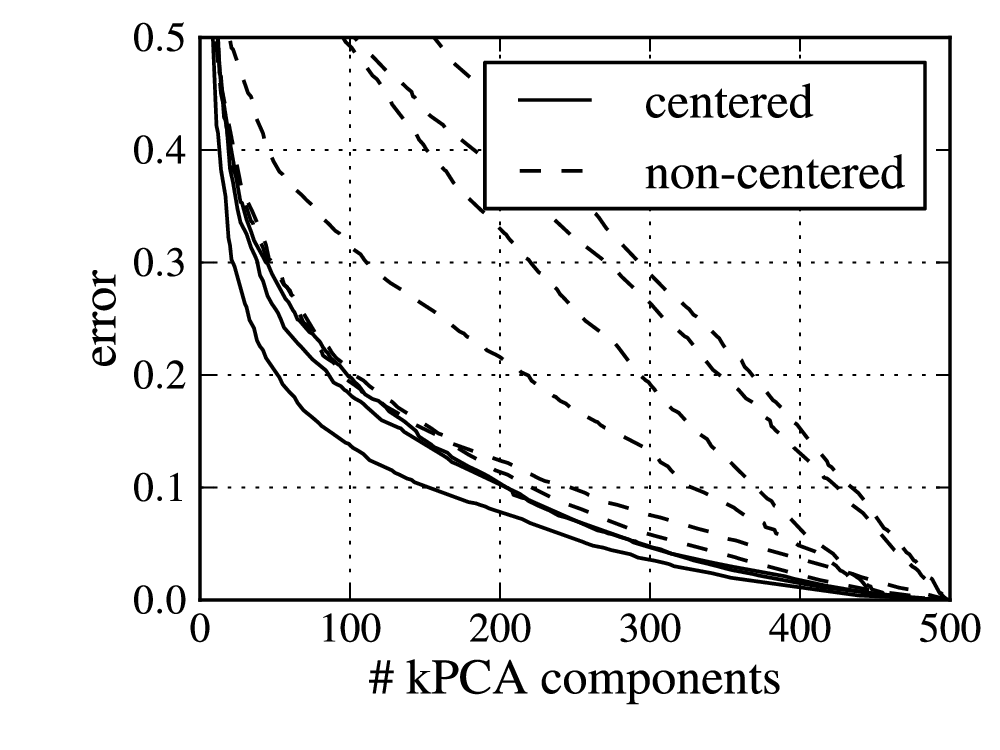} &
\includegraphics[scale=0.7,trim=10 0 0 0]{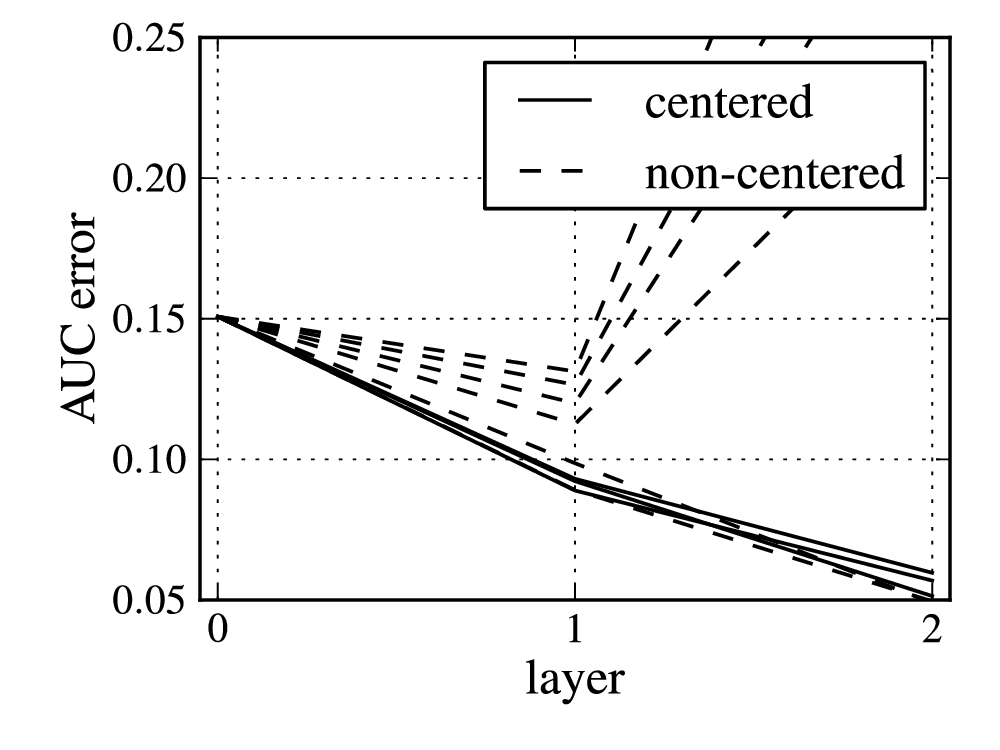}
\end{tabular}\vskip -5mm
\caption{On the left, residuals of the projection of the labels in the leading components of the top layer kernel $k_2(x,x')$ (see Equation \ref{equation-kpca}) after 10 epochs. On the right, layer-wise evolution of the representation in terms of area under the error curve (see Equation \ref{equation-auc}) after 100 epochs. Centered DBMs are more stable than non-centered ones. Top layer representations are clearly better than the input.}
\label{figure-plotkpca}
\vskip 2cm

\centering
\begin{tabular}{cc}
~~~~Discriminative performance & ~~~~Generative performance\\
\includegraphics[scale=0.7,trim=10 0 0 0]{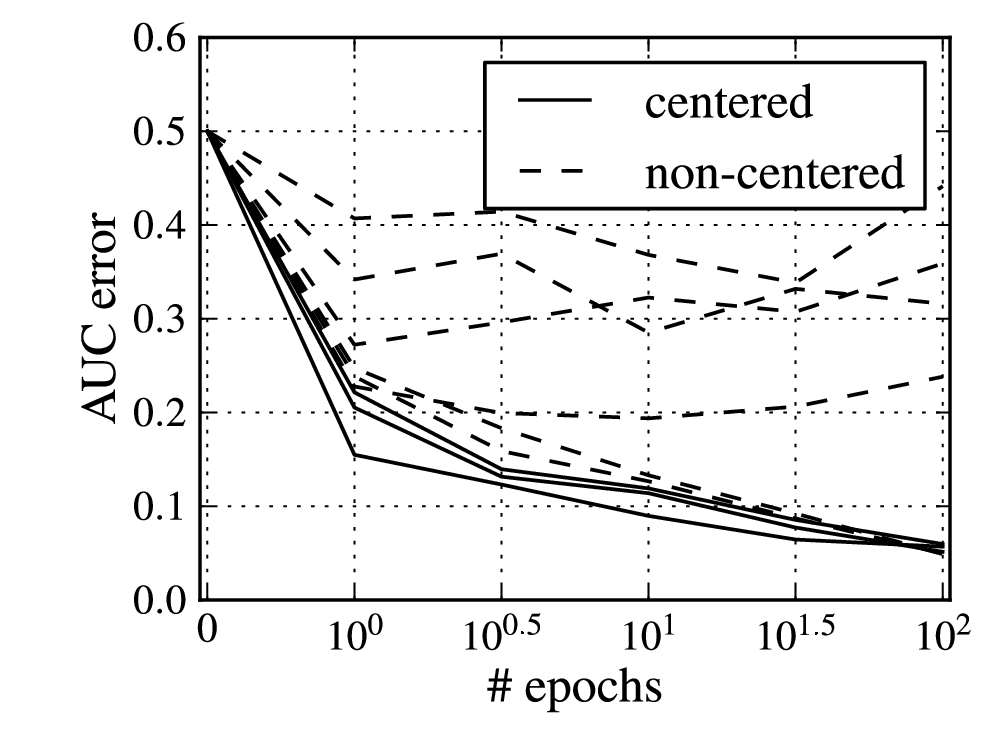} &
\includegraphics[scale=0.7,trim=10 0 0 0]{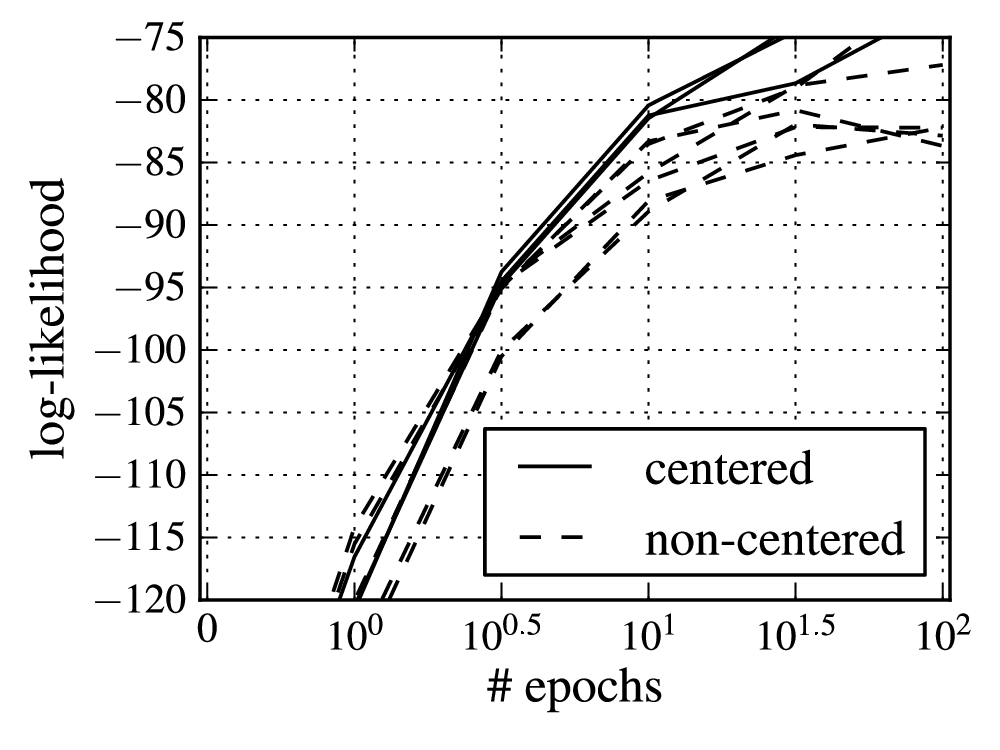}
\end{tabular}\vskip -5mm
\caption{Convergence speed of centered and non-centered DBMs (in terms of top layer AUC error and model log-likelihood). Centered DBMs learn faster and are more stable than non-centered ones. Note that the estimate of the log-likelihood from Equation \ref{equation-ais-loglikelihood} becomes inaccurate as the model becomes more complex (after 10 epochs).}
\label{figure-plotconvergence}
\end{figure}

Table \ref{table-discriminative} corroborates the importance of centering for better discriminating in the top layer of a deep Boltzmann machine.
As it can be seen in Figure \ref{figure-plotconvergence} (left), discriminative performance of the top layer can be further improved by training the network for a longer time.

Table \ref{table-generative} further supports the importance of centering, showing that centered DBMs learn a better generative model of data. However, the advantage is not as strong as for the discriminative case. Indeed, units in the top layer are not critical for generative performance as the learning algorithm can simply discard them and learn a one-layer shallow generative model instead.

Figure \ref{figure-plotkpca} and \ref{figure-plotconvergence} highlight the importance of centering for faster and more stable learning. The models emerging from the centered deep Boltzmann machine have systematically better discriminative properties in the top layer and good generative properties. While a non-centered DBM may ultimately learn a model which is as good as the one produced by a centered DBM, it may also diverge.

\begin{table}
\centering
\begin{tabular}{l|ccc}
AUC error & $b_0,c_0 = 2$  &$b_0,c_0 = 0$  &$b_0,c_0 = -2$\\[1pt]\hline
& & & \\[-12pt]
$\beta,\gamma = \sigm(2)$ & \textbf{0.119} & 0.194 & 0.285 \\
$\beta,\gamma = \sigm(0)$ & 0.133 & \textbf{0.090} &  0.127 \\
$\beta,\gamma = \sigm(-2)$  & 0.368 & 0.323 & \textbf{0.114}
\end{tabular}
\caption{Discriminative performance after 10 epochs in the top layer of the deep Boltzmann machine as measured by Equation \ref{equation-auc} for different configurations of initial bias and offset. The lower the AUC error the better. In each case, centering sigmoids  leads to better discrimination in the top layer.}
\label{table-discriminative}
\end{table}

\begin{table}
\centering
\begin{tabular}{l|ccc}
$\langle \log p(x;\theta) \rangle_\text{data}$& $b_0,c_0 = 2$  & $b_0,c_0 = 0$ & $b_0,c_0 = -2$\\[1pt]\hline
& & & \\[-12pt]
$\beta,\gamma = \sigm(2)$ & \textbf{-81.5}\footnotemark &-86.5\footnotemark[\value{footnote}] & -88.9\footnotemark[\value{footnote}] \\
$\beta,\gamma = \sigm(0)$ & -83.5\footnotemark[\value{footnote}]  & \textbf{-81.2}\footnotemark[\value{footnote}] & -85.6\footnotemark[\value{footnote}]  \\
$\beta,\gamma = \sigm(-2)$ & -88.1\footnotemark[\value{footnote}] & -83.3\footnotemark[\value{footnote}] & \textbf{-80.4}\footnotemark[\value{footnote}]
\end{tabular}
\caption{Generative performance after 10 epochs in terms of estimated model log-likelihood $\langle \log p(x;\theta) \rangle_\text{data}$ for different configurations of initial bias and offset. The generative performance is less sensitive to the initial conditioning of the DBM than the top layer discriminative performance as the top-level units can simply be discarded, leading essentially to a more robust one-layer generative model.}
\label{table-generative}
\end{table}\footnotetext{In some other research work, authors are computing a lower bound of the log probablity instead of a direct estimate of it, thus making a direct comparison impossible. Also, estimates of log probability become increasingly inaccurate as the model $\theta$ complexifies.}

\newcolumntype{C}{>{\centering\arraybackslash} m{3.5cm} }

\newcommand{\filt}[1]{\includegraphics[scale=0.6]{m-#1.eps}}
\newcommand{\filtpage}[2]{%
\begin{tabular}{m{2.75cm}CCC}
& $b_0,c_0 = 2$ & $b_0,c_0 =0$ & $b_0,c_0 =-2$\\[5pt]
$\beta,\gamma = \sigm(2)$ & \filt{a-a-#1-#2}&\filt{c-a-#1-#2}&\filt{s-a-#1-#2}\\
$\beta,\gamma = \sigm(0)$ & \filt{a-c-#1-#2}&\filt{c-c-#1-#2}&\filt{s-c-#1-#2}\\
$\beta,\gamma = \sigm(-2)$ & \filt{a-s-#1-#2}&\filt{c-s-#1-#2}&\filt{s-s-#1-#2}\\
\end{tabular}
}
\begin{figure}
\centering
\filtpage{4}{W}
\caption{Examples of first-layer filters of the DBM for different bias and offset parameters after 100 epochs. These filters are rendered using a linear backprojection of top layer units onto the input space. Each model is producing reasonable first-layer filters, suggesting that one-layer networks (i.e. restricted Boltzmann machines) are less sensitive to the quality of the conditioning of the parameter space.}
\label{figure-filtersW}
\vskip 1cm
\filtpage{4}{V}
\caption{Examples of second-layer filters of the DBM for different bias and offset parameters after 100 epochs. These filters are rendered using a linear backprojection of intermediate layer units onto the input space. Here, we can clearly see that the diversity of filters is higher when the DBM is centered.}
\label{figure-filtersV}
\vskip 1cm
\filtpage{2}{X}
\caption{Examples of digits generated by the DBM for different bias and offset parameters after 10 epochs. The degenerated second layer of the non-centered DBM seems to have a negative impact on the balance between different classes.}
\label{figure-filtersX}
\end{figure}

\newcommand{\pcaplot}[1]{\includegraphics[scale=0.46,trim=5 5 5 5,clip=true]{m-#1-0-l2.eps}}
\newcommand{\pcaplotlate}[1]{\includegraphics[scale=0.46,trim=5 5 5 5,clip=true]{m-#1-4-l2.eps}}
\begin{figure}
\centering \small
\hrule \vskip 1mm Top layer representation after $1$ epoch \vskip 1mm \hrule \vskip 2mm
\begin{tabular}{m{2.5cm}CCC}
& $b_0,c_0 = 2$ & $b_0,c_0 =0$ & $b_0,c_0 =-2$\\[3pt]
$\beta,\gamma = \sigm(2)$ & \pcaplot{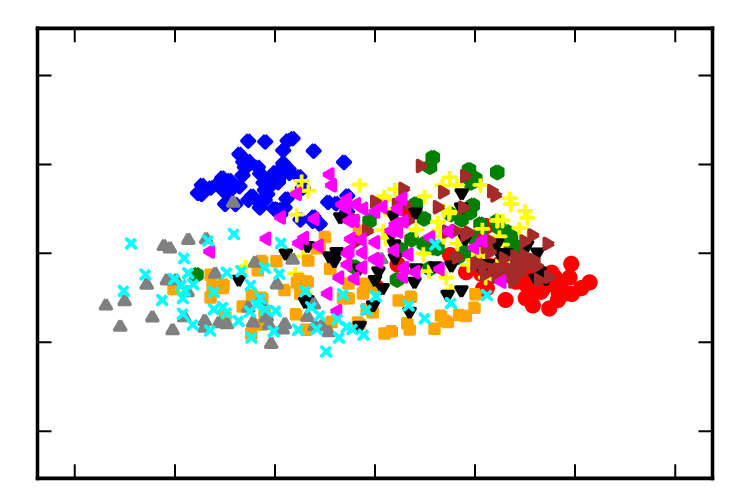}&\pcaplot{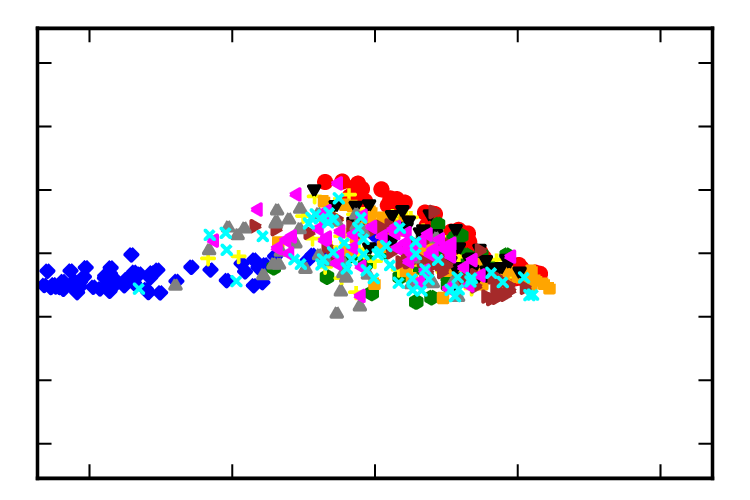}&\pcaplot{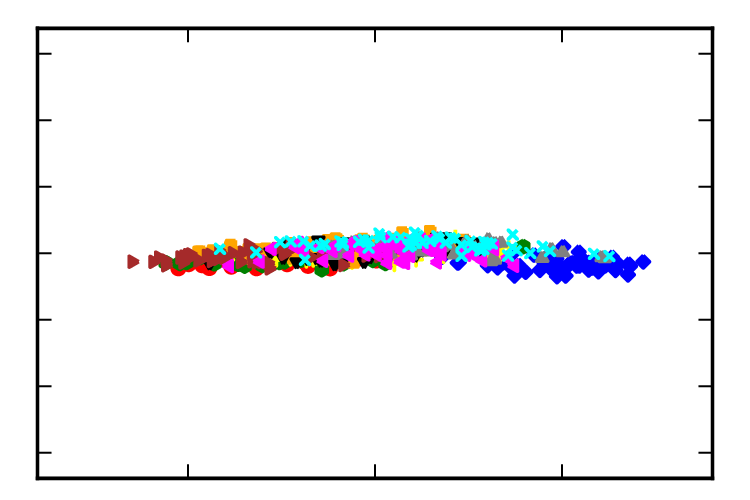}\\
$\beta,\gamma = \sigm(0)$ & \pcaplot{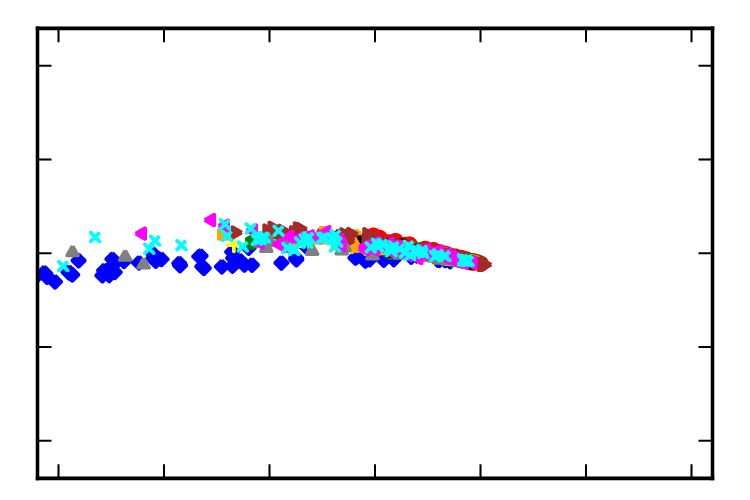}&\pcaplot{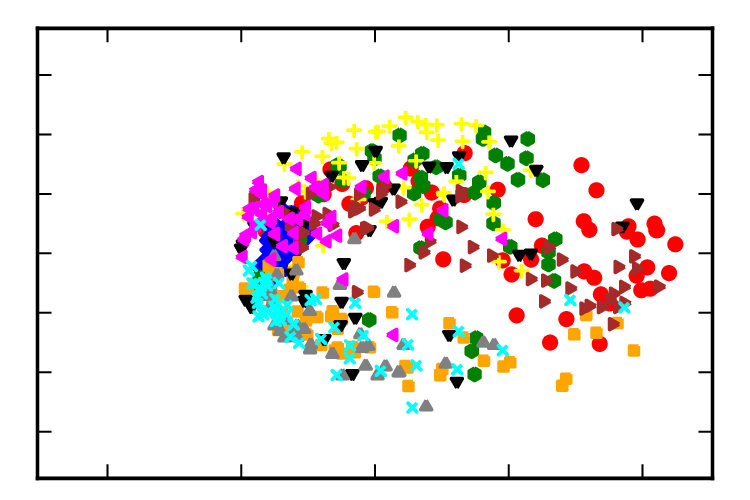}&\pcaplot{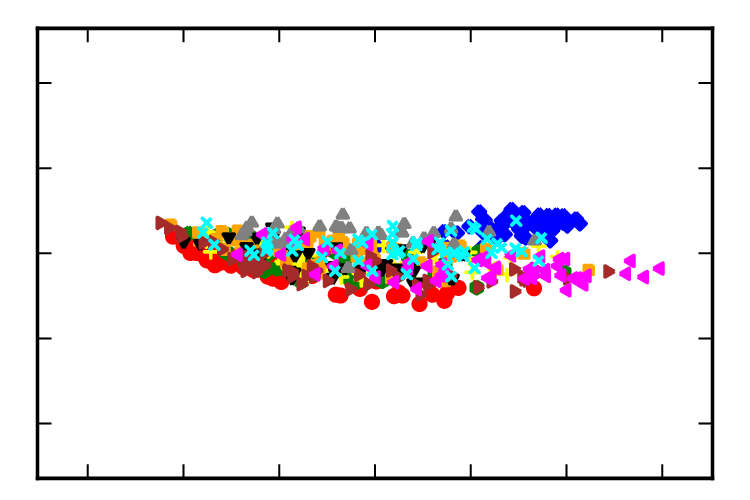}\\
$\beta,\gamma = \sigm(-2)$ & \pcaplot{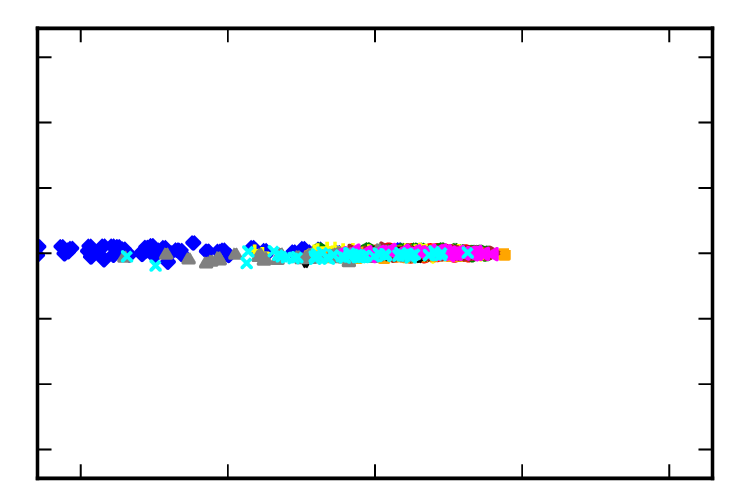}&\pcaplot{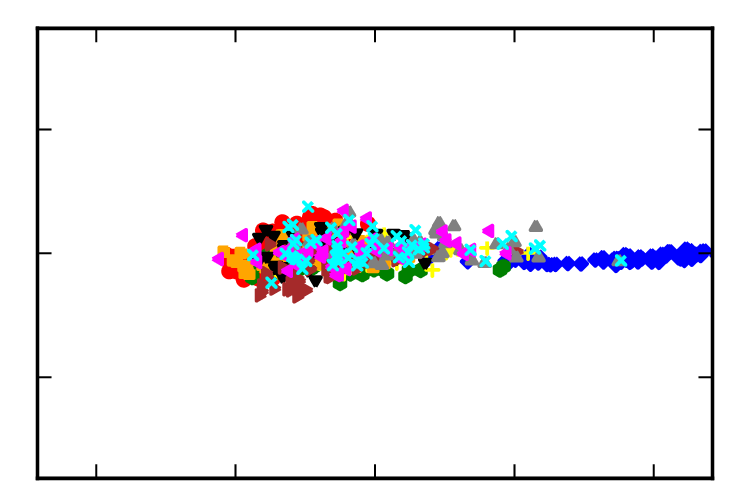}&\pcaplot{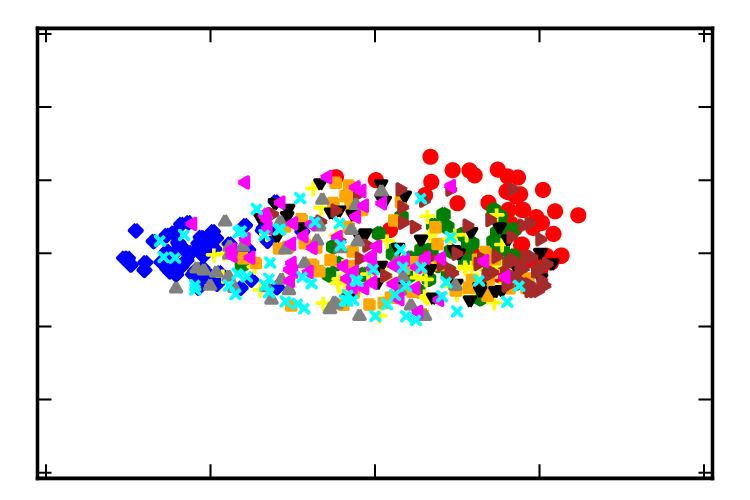}
\end{tabular}
\vskip 5mm
\hrule \vskip 1mm Top layer representation after $100$ epochs \vskip 1mm \hrule \vskip 2mm
\begin{tabular}{m{2.5cm}CCC}
& $b_0,c_0 = 2$ & $b_0,c_0 =0$ & $b_0,c_0 =-2$\\[3pt]
$\beta,\gamma = \sigm(2)$ & \pcaplotlate{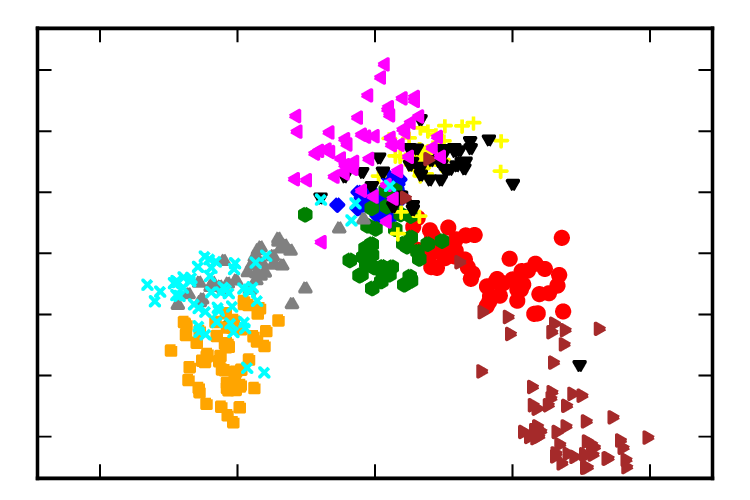}&\pcaplotlate{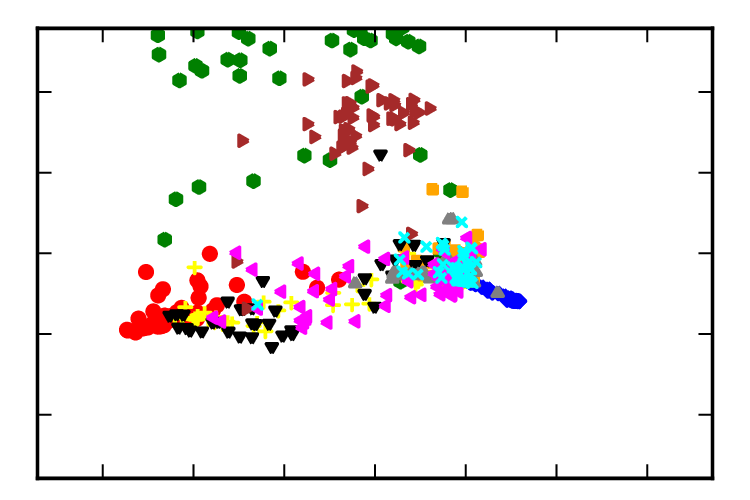}&\pcaplotlate{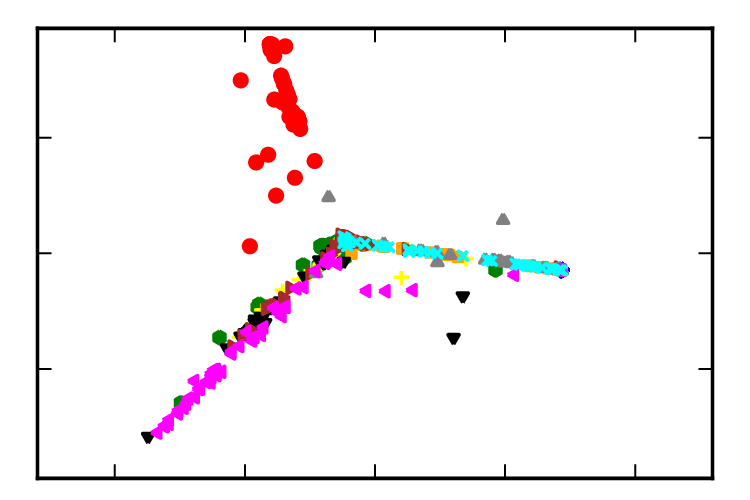}\\
$\beta,\gamma = \sigm(0)$ & \pcaplotlate{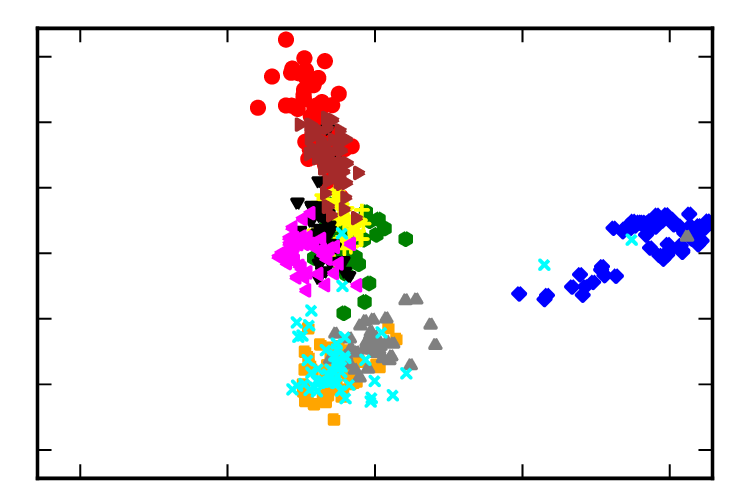}&\pcaplotlate{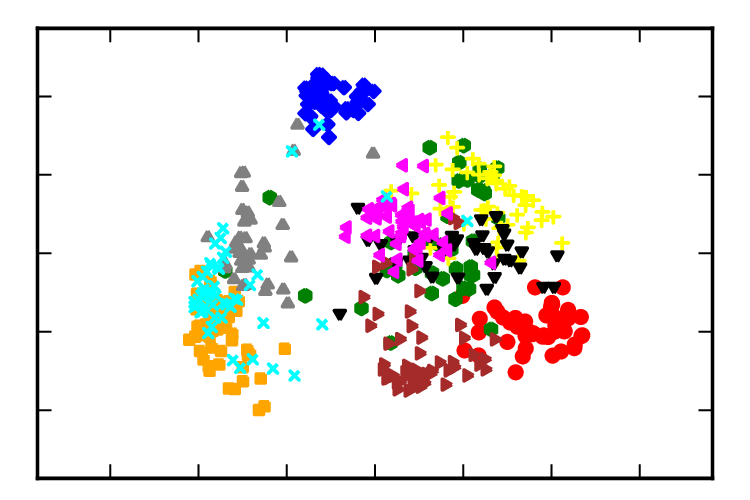}&\pcaplotlate{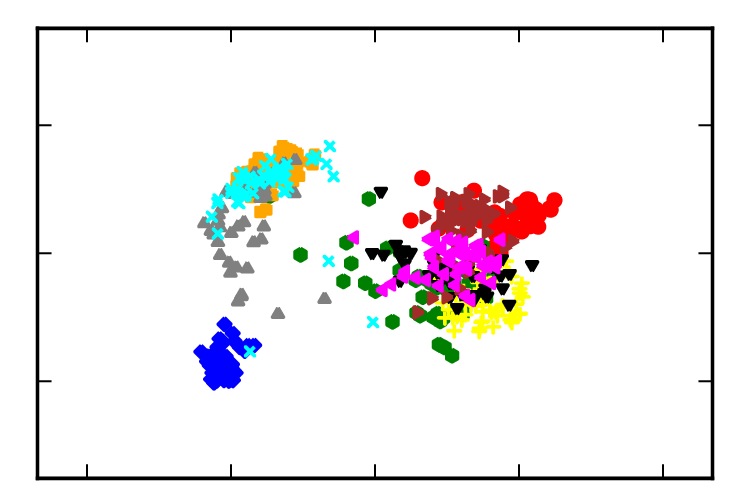}\\
$\beta,\gamma = \sigm(-2)$ & \pcaplotlate{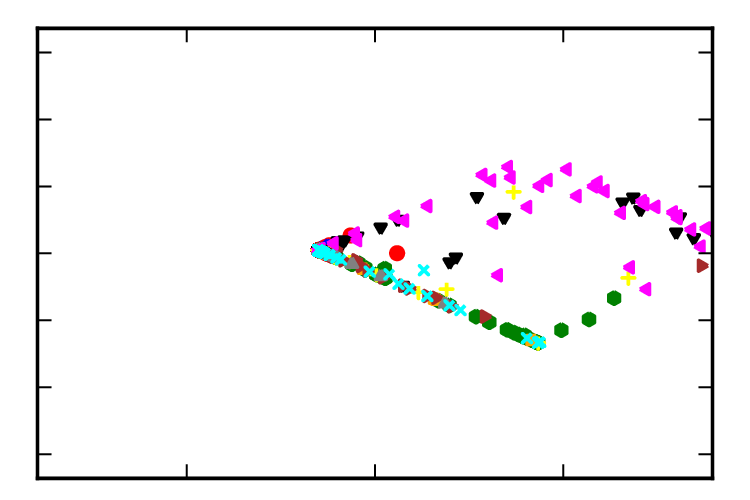}&\pcaplotlate{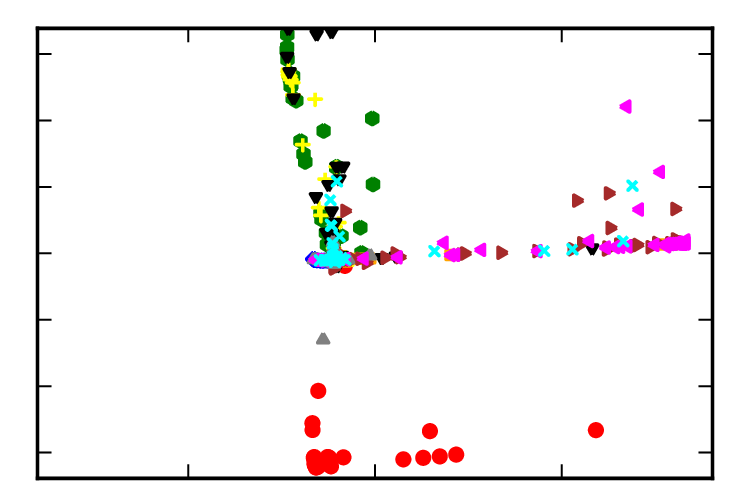}&\pcaplotlate{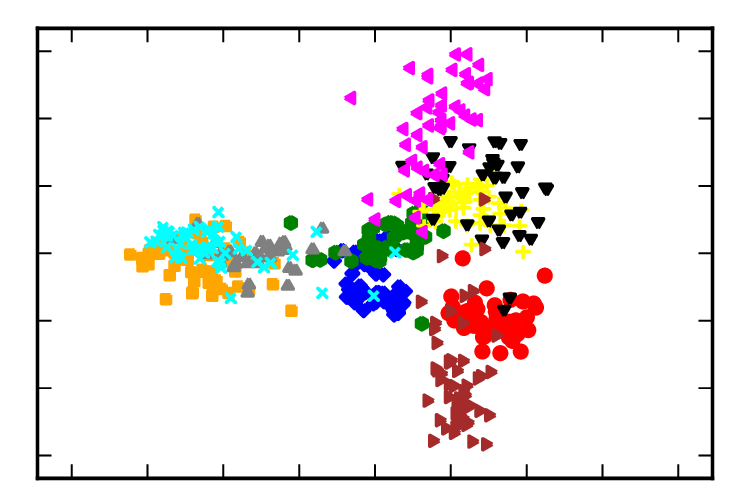}
\end{tabular}
\caption{2-kPCA visualization of the top-level representation in the DBM for different bias and offset parameters at different stages of training. Points are colored according to their label (``0''=red, ``1''=blue, ``2''=green, ``3''=yellow, ``4''=orange, ``5''=black, ``6''=brown, ``7''=gray, ``8''=magenta, ``9''=cyan). Non-centered DBMs tend to collapse the data onto a simplistic low-dimensional manifold in the top layer representation. On the other other hand, in the centered DBM, we can clearly observe in the late stage of training the emergence of clusters corresponding to labels.}
\label{figure-2pca}
\end{figure}

Figure \ref{figure-filtersW} and \ref{figure-filtersV} show that each model is able to learn reasonable first-layer filters but that second-layer filters learned by a centered DBM tend to be more varied than those learned by a non-centered DBM. This higher variety of second layer filters suggests that the centered DBM produces a richer top-level representation. The argument is corroborated by Figure \ref{figure-2pca} showing that, in absence of centering mechanism, the projection of the data on the top layer representation tends to form a simplistic low-dimensional manifold that may still contain useful features (for example, discriminating the digit ``1'' from other digits) but, on the other hand, that also discards a lot of potentially useful discriminative features. As suggested by Figure \ref{figure-filtersX}, the top-layer simplistic representation may even negatively affect the generative properties of the model by perturbing the balance between different classes.

\section{Conclusion}

We presented a simple modification of the deep Boltzmann machine that centers the output of the sigmoids by rewriting the energy function as a function of centered states. This centered version of the deep Boltzmann machine is easy to implement as it simply involves a reparameterization of the energy function. A theoretical motivation for centering is that it leads to a better conditioning of the Hessian of the optimization criterion.

This simple modification allows to learn efficiently a deep Boltzmann machine without greedy layer-wise pretraining. Experiments on real data corroborate the benefits of centering, showing that the centered deep Boltzmann machine learns faster and is more stable than its non-centered counterpart. In addition, the centered deep Boltzmann machine produces useful discriminative features in the top layer and a good generative model of data.

Training hierarchies of many layers is still tedious and requires many iterations. Understanding whether the difficulty comes from a difficult optimization problem or from the exhaustion of statistical information in the data set remains to be done. Also, despite an initial good conditioning of the Hessian, it can not be excluded that the solution progressively drifts towards degenerate regions of the parameter space throughout the learning procedure. Strategies to dynamically maintain the solution within well-behaved regions of the parameter space or to better descend the objective function also need to be further investigated.

\acks{This work was supported by the World Class University Program through the National Research Foundation of Korea funded by the Ministry of Education, Science, and Technology, under Grant R31-10008.}

\bibliography{paper}

\end{document}